\documentclass[10pt,twocolumn,letterpaper]{article}

\usepackage{cvpr}              
\definecolor{cvprblue}{rgb}{0.21,0.49,0.74}
\usepackage[pagebackref,breaklinks,colorlinks,allcolors=cvprblue]{hyperref}
\usepackage{CJKutf8}
\usepackage{multirow}

\def\paperID{6} 
\def\confName{CVPRW}
\def\confYear{2026}

\title{Addressing Image Authenticity When Cameras Use Generative AI}

\author{
\setlength{\tabcolsep}{20pt}
\begin{tabular}{ccc}
Umar Masud$^1$\thanks{Work done during an internship at Samsung.} & Abhijith Punnappurath$^2$ & Luxi Zhao$^2$\thanks{Work done while with Samsung.} \\
{\tt\small um71000@gmail.com} & {\tt\small abhijith.p@samsung.com} & {\tt\small lucyzhao.zlx@gmail.com} \\[1ex]
\multicolumn{3}{c}{David B. Lindell$^1$ \hspace{2cm} Michael S. Brown$^2$} \\
\multicolumn{3}{c}{{\tt\small lindell@cs.toronto.edu} \hspace{1cm} {\tt\small michael.b1@samsung.com}} \\[1ex]
\multicolumn{3}{c}{$^1$University of Toronto \quad \quad $^2$AI Center--Toronto, Samsung Electronics.}
\end{tabular}
}

\begin{document}
\maketitle
\begin{abstract}
The ability of generative AI (GenAI) methods to photorealistically alter camera images has raised awareness about the authenticity of images shared online. Interestingly, images captured directly by our cameras are considered authentic and faithful. However, with the increasing integration of deep-learning modules into cameras' capture-time hardware—namely, the image signal processor (ISP)—there is now a potential for hallucinated content in images directly output by our cameras. Hallucinated capture-time image content is typically benign, such as enhanced edges or texture, but in certain operations, such as AI-based digital zoom or low-light image enhancement, hallucinations can potentially alter the semantics and interpretation of the image content. As a result, users may not realize that the content in their camera images is not authentic. This paper addresses this issue by enabling users to recover the ``unhallucinated" version of the camera image to avoid misinterpretation of the image content. Our approach works by optimizing an image-specific multi-layer perceptron (MLP) decoder together with a modality-specific encoder so that, given the camera image, we can recover the image before hallucinated content was added. The encoder and MLP are self-contained and can be applied post-capture to the image without requiring access to the camera ISP. Moreover, the encoder and MLP decoder require only 180 KB of storage and can be readily saved as metadata within standard image formats such as JPEG and HEIC.
\end{abstract}

\section{Introduction}
\label{sec:intro}

\begin{figure}[t]
\centering
\includegraphics[width=\linewidth]{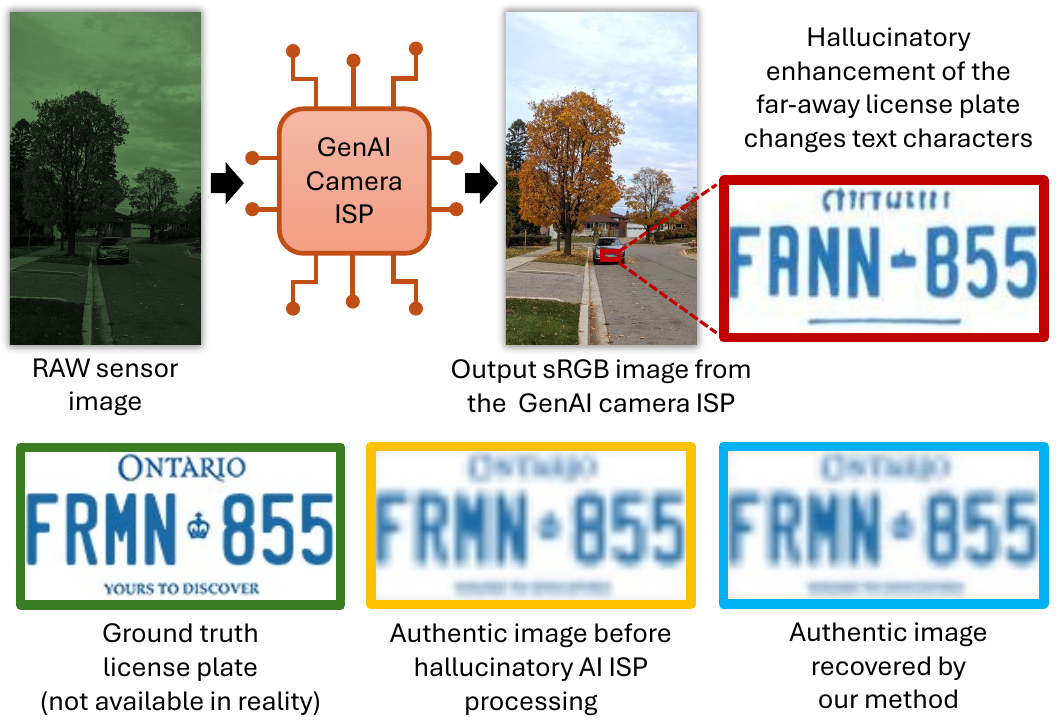}
\caption{Modern smartphone image signal processors (ISPs) are starting to incorporate GenAI modules that can hallucinate scene content for specific imaging modalities (e.g., digital zoom and low-light/night photography). For example, a license plate that the camera's native optics could not resolve might be enhanced by GenAI-based digital zoom. In this case, the letters `R' and `M' have been hallucinated as `A' and `N', respectively, and the digit `8' looks like a `B'. To an unaware user, this camera image would be considered authentic and would avoid scrutiny. The problem we are interested in addressing is how to reveal to the user the \textit{unhallucinated} or authentic version of the image before any hallucinatory enhancements were introduced by the camera ISP.
\label{fig:teaser}
}
\end{figure} 

\begin{figure*}[t]
\centering
\includegraphics[width=\linewidth]{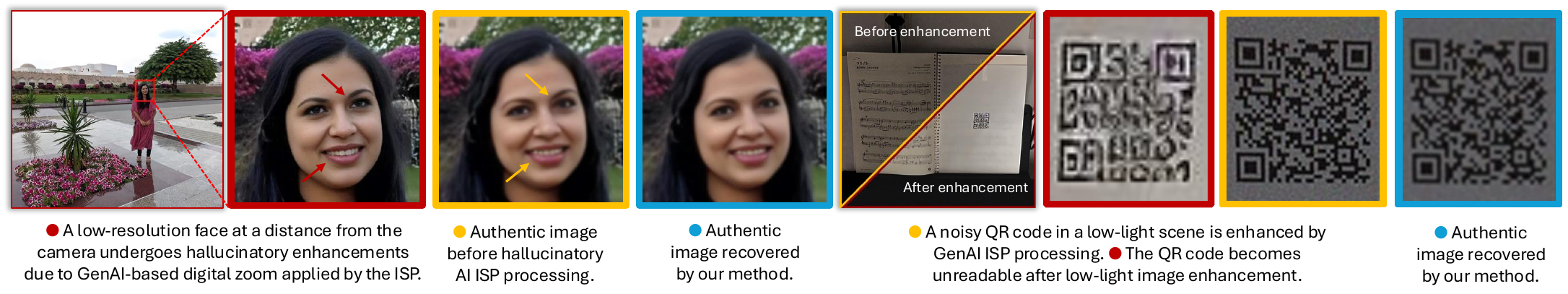}
\caption{GenAI-based super-resolution applied by the ISP to faces can sometimes result in subtle changes to appearance or even identity. Notice the change in gaze and the shape of the mouth and teeth. GenAI-based low-light image enhancement is another ISP operation that can potentially alter the information conveyed by an image. The enhanced image is aesthetically more appealing. However, the QR code which reveals a URL when scanned becomes unreadable post enhancement. Our method allows the user to recover the authentic face and a readable QR code.
\label{fig:extended_teaser} 
}
\end{figure*} 

Until recently, producing altered or \textit{deepfake} images required specialized software and computing resources not accessible to the majority of users.  In the last few years, however, powerful generative AI (GenAI) tools have become readily available that make it possible to produce photorealistic image alterations with just a few clicks or text prompts (e.g.,~\cite{df1,faceapp,firefly,midjourney,dalle3}). The implications of such GenAI to alter and create images are profound. Fakes and altered images circulated at scale can, for example, sway political sentiment, influence election outcomes, or create confusion and spread misinformation about serious issues~\cite{agarwal2019protecting,dobber2021microtargeted,hameleers2022you,vaccari2020deepfakes}.  This has prompted governments and regulatory bodies to introduce legislation to curb GenAI misuse~\cite{eu_act}. Rules are being introduced requiring developers and deployers to ensure end-users are aware that they are interacting with AI imagery. Major social media platforms have already begun tagging and prominently demarcating photos that were edited or born digital using generative AI methods~\cite{fb_news,google_news}. Awareness of deepfakes and GenAI-altered images is now mainstream.

Discussions about altered camera images remain focused on \textit{post-capture} manipulation, assuming images directly output by cameras as authentic and trusted renditions of the physical scene. This is no longer true~\cite{access}. Cameras employ dedicated onboard hardware called an image signal processor (ISP), which applies various processing steps to the sensor response---such as denoising, demosaicing, auto white balance, color and tonal adjustments, sharpening, digital zoom, and more---to produce the photo displayed to the user~\cite{MobileTour,ISPBrown}. Traditionally, signal-processing algorithms have served as the underpinning of ISP processing blocks. While conventional signal-processing-based methods often manipulate the colors and tones of an image, they are not designed to generate ``hallucinated'' image content. However, modern ISPs increasingly use AI-based modules, especially for challenging imaging scenarios such as digital zoom and night photography. When AI models trained with generative or perceptual losses are used in ISPs, they are prone to hallucinate content, potentially altering image meaning~\cite{access}. The implication is that images directly output from the camera may now contain ``fake'' content, especially in smartphone cameras where AI-ISP modules are seeing increasing adoption. The use of GenAI in camera hardware marks a paradigm shift in how we view camera images and challenges the traditional forensic view of camera-captured images as inherently trustworthy.

Currently, most hallucinated content from AI-based ISP modules is arguably benign---hallucinations manifest as sharper edges or textures that do not change the interpretation of the image. However, certain ISP stages, in particular when using GenAI models, can alter the semantics and the interpretation of the image. The most prominent operation in this category is super-resolution (SR) or digital zoom. Unlike DSLRs, smartphone cameras have limited optical zoom capabilities owing to their small form factor. AI-based super-resolution is commonly used to achieve higher zoom factors, with an increasing probability of hallucinated content with increasing zoom. Fig.~\ref{fig:teaser} shows a license plate and the first example in Fig.~\ref{fig:extended_teaser} shows a face enhanced by a GenAI zoom module. Another operation is GenAI-based low-light image enhancement, where the challenging situation of high noise and lack of detail in dark regions leads to hallucinated content (see the second example in Fig.~\ref{fig:extended_teaser}). 

To our knowledge, the only work to address the authenticity of camera-captured images is that of Punnappurath et al.~\cite{access}, which proposed a binary authentication mask be computed at capture time and stored in the camera image as metadata.  The binary mask reveals to the user which pixels in the image are considered potentially hallucinated. This approach, however, cannot reveal what the \textit{unhallucinated} or authentic image would have looked like. This may be of interest in specific scenarios, such as those shown in Figs.~\ref{fig:teaser} and~\ref{fig:extended_teaser}. The ability to recover an unhallucinated version of the camera image serves as the impetus of our work.

\noindent \textbf{Contribution}~
We present a method that enables users to recover an unhallucinated or authentic version of a camera-captured image, effectively reversing hallucinatory AI-based ISP processing as if no such modifications had been applied. Our method is metadata-assisted and is based on implicit neural representations, which have shown strong performance across tasks such as image compression~\cite{strumpler2022implicit}, 3D representation~\cite{nerf}, and SR~\cite{gao2023implicit}. However, unlike existing coordinate-based multi-layer perceptron (MLP) approaches, such as SIREN~\cite{siren} or NeRF~\cite{nerf}, which are optimized per image and converge slowly, we propose a lightweight encoder and MLP decoder architecture. The encoder generates a learned embedding that, combined with the image pixel's $(x,y)$ coordinates, serves as input to the MLP decoder.
The encoder is pretrained per modality (for each class of hallucinatory ISP operation) and remains fixed, while the MLP decoder can be rapidly finetuned per image at capture time. Our metadata consists of the encoder and MLP parameters, resulting in a compact model size of only 180 KB, which can be easily stored within standard image formats such as JPEG or HEIC.  Our solution operates independently of the AI module responsible for hallucinations, assuming no prior knowledge of this model and requiring no post-capture access to the camera ISP.

\section{Related work}
\label{sec:related}

\textbf{Digital image forensics and neural ISPs.} Image forensics provides tools to validate whether digital images are authentic or fake~\cite{farid2009image,popescu2004statistical,KORUS20171}. Forensic approaches can broadly be classified into passive and active methods. Passive methods operate under the assumption that tampering alters the underlying statistics of the image, and rely on anomalies in the image itself for authentication. 
On the other hand, active methods rely on additional information accompanying the image as a digital signature or watermark and are typically more accurate than their passive counterparts. At its simplest, authentication seeks to detect whether or not an image has been altered. More advanced methods allow localization of the manipulated regions~\cite{zhang2024editguard,guillaro2023trufor}, or even approximately restoring the affected pixels~\cite{ying2023learning,recovery_access}. However, such methods are limited to the domain of post-capture image forensics and not ISP-induced hallucinations itself.

The rise of AI-generated imagery has mounted new challenges to image forensics. There is a growing body of research around detecting deepfakes and images born digitally~\cite{farid2022creating,nguyen2022deep,verdoliva2020media}. Concurrently, researchers have recently begun exploring the feasibility of replacing conventional signal-processing-based ISP modules with neural modules to improve camera image quality, particularly on smartphones~\cite{ershov2022ntire,shutova2023ntire,ignatov2020aim,ignatov2019aim,ignatov2021learned,ignatov2022learned}. These neural ISPs have started to achieve near-real-time performance on smartphone devices~\cite{ignatov2021learned,ignatov2022learned} leading to their increased adoption on camera ISP hardware. Many of these neural ISP modules share the same characteristics as the generative AI models used for post-capture image editing in terms of the hallucinatory loss functions (e.g., VGG~\cite{johnson2016perceptual}, LPIPS~\cite{zhang2018perceptual}, GAN~\cite{goodfellow2014generative}) used for training and their tendency to generate fake image content. While the issue this creates of \textit{on-camera} hallucinations has only very recently been addressed in~\cite{access}, a small number of works have proposed to leverage neural imaging pipelines to improve the detection and localization accuracy of \textit{post-capture} manipulations~\cite{hu2023draw,korus2019content}. These methods embed a digital watermark into the RAW image~\cite{hu2023draw} or modify the neural ISP training objective with authentication losses~\cite{korus2019content} so that post-capture forgery of the sRGB image can be robustly detected and localized. However, \cite{hu2023draw,korus2019content} do not discuss the issue of non-authentic content generated by GenAI processing on the ISP itself. Furthermore, they are limited to detection and localization of the manipulated pixels. In contrast, our focus is on hallucinations produced by AI modules onboard the ISP, and we target the full recovery of the authentic image.

\begin{figure*}[t]
\centering
\includegraphics[width=0.9\linewidth]{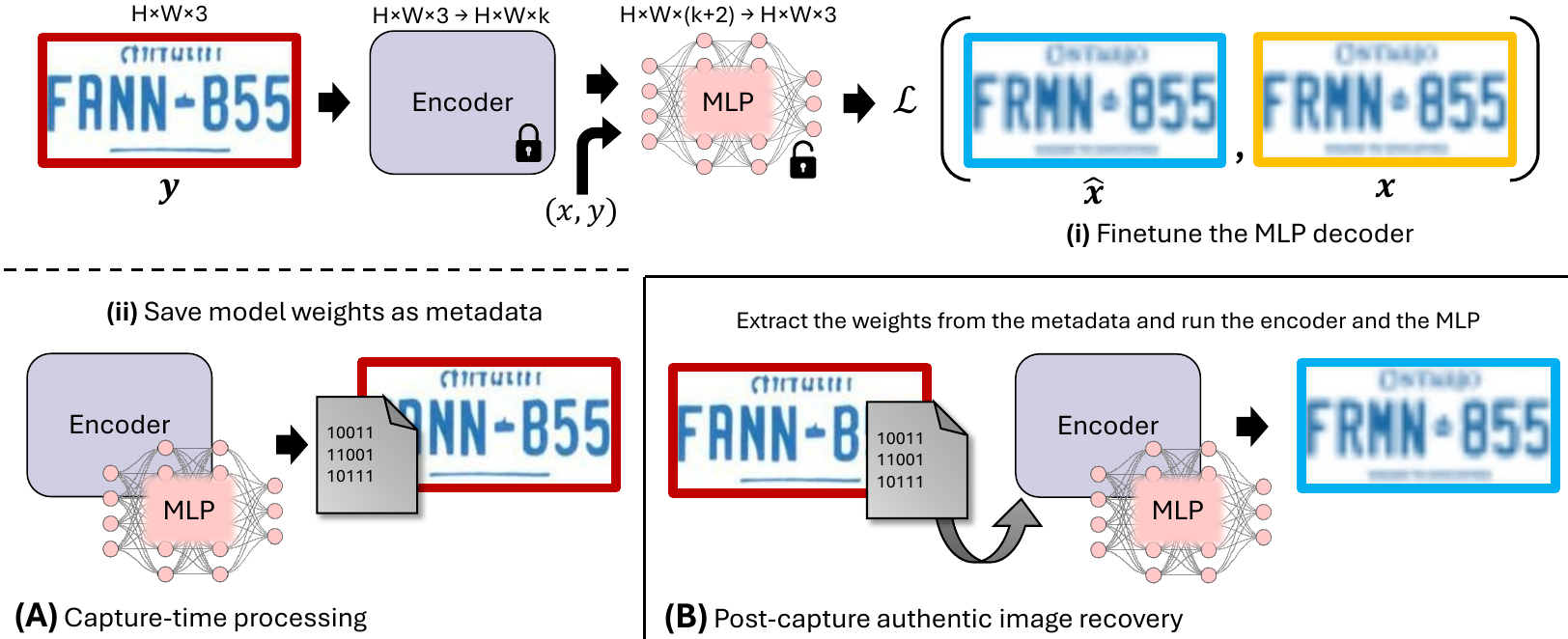}
\caption{An overview of our proposed method. (A) At capture time, we (i) run the ISP's output image $\mathbf{y}$ containing the hallucinations through the frozen pretrained encoder. The $(x,y)$ coordinates are concatenated with the encoder latents and provided as input to the MLP decoder for finetuning. The MLP decoder operates per-pixel: for each spatial location $(x,y)$, the corresponding $k$-dimensional encoder feature vector is concatenated with the normalized $(x,y)$ coordinates and passed through the MLP to obtain its prediction at that pixel. The $H\!\times\!W\!\times\!(k+2) \rightarrow {H\!\times\!W\!\times\!3}$ notation denotes this per-pixel operation applied across all spatial locations. The MLP performs residual prediction to recover the unhallucinated image $\mathbf{\hat{x}}$ and the loss is computed against the ground truth authentic image $\mathbf{x}$ before hallucinations were introduced by AI ISP processing. (ii) Once finetuned, the encoder and the MLP model parameters are saved as metadata along with the hallucinated image. (B) At inference, the model weights are extracted from the metadata and the hallucinated image is run through the encoder and the MLP to obtain the unhallucinated image.
\label{fig:method}}
\end{figure*}

While an entire camera imaging pipeline can be replaced with a single monolithic neural model, a modular design is more practical as it offers better control and interpretability~\cite{twostageisp,deepflexisp}. In CameraNet~\cite{twostageisp}, the pipeline is divided into two sub-networks: Restore-Net and Enhance-Net. Restore-Net, a RAW front-end network, performs denoising, demosaicing, and white balance to restore the underlying signal. Enhance-Net, a photofinishing back-end, applies tone and color adjustments and other enhancements to improve aesthetic quality. Similarly, DeepFlexISP~\cite{deepflexisp}, the winner of the night photography rendering challenge in~\cite{ershov2022ntire}, partitions the camera pipeline into three sub-networks for denoising, white balance, and photofinishing. The denoiser is trained with a pure reconstruction loss, while the other two networks incorporate enhancement losses. As noted in~\cite{access}, perceptual and generative losses used for image enhancement often introduce fake content. These losses are typically applied to neural enhancement modules at the end of the ISP pipeline, particularly in operations such as super-resolution and low-light image enhancement.

\noindent \textbf{Super-resolution and low-light image enhancement.}
Since the seminal work in SRCNN~\cite{srcnn}, which introduced deep learning for the SR task, many of the improvements in image quality have been driven by moving away from reconstruction losses, which typically produce blurry results, towards perceptual and generative losses. 
SR models based on both GANs~\cite{wang2021realesrgan,ledig2017photo} and diffusion~\cite{wang2024sinsr,yue2024resshift} have achieved remarkable success in recent years.
These models produce sharper outputs with a higher perceptual image quality often by hallucinating fake details and texture. The best performing methods for text SR also use such generative loss functions~\cite{li2023marconet,zhang2024diffusion}. At the same time, it is becoming increasingly feasible to run such GenAI SR solutions in real time on smartphone hardware~\cite{ignatov2022efficient}. Similarly, models using perceptual and generative losses dominated the recent challenges on low-light image enhancement and nighttime photography~\cite{ershov2022ntire,shutova2023ntire}.

\noindent \textbf{Implicit neural representations.} 
Coordinate-based implicit neural representations, such as SIREN~\cite{siren} and NeRF~\cite{nerf}, reconstruct images using an MLP that takes $(x,y)$ coordinates as input. While these methods achieve excellent performance, they are typically optimized per image and converge slowly.  Some approaches, such as hashgrid~\cite{hashgrid}, trade off model size for faster optimization. Instead of feeding coordinates directly, they use a learnable encoding---such as a hashtable in~\cite{hashgrid}--which can be quickly optimized before being input to the MLP. While this trade-off is effective for gigapixel images and 3D tasks, it is suboptimal for normal-resolution images. We propose a learnable encoding using a lightweight neural encoder, which we demonstrate in our experiments to be a more effective choice than other MLP variants.

\noindent \textbf{Metadata-assisted image recovery and MLPs.} Capture-time metadata has been shown to significantly improve various image recovery tasks. RAW reconstruction methods~\cite{rang,Nam_2022_CVPR,li2023metadata,wang2023raw} store a small amount of metadata alongside the sRGB image at acquisition time. Similarly, GamutMLP~\cite{le2023gamutmlp} introduces a metadata-assisted gamut recovery approach. MLPs are well-established as compact image representations, particularly in image compression~\cite{strumpler2022implicit}. Inspired by GamutMLP, our method optimizes a model and stores its parameters as image metadata. However, we demonstrate that a lightweight encoder paired with an MLP decoder outperforms a standalone MLP for the authentication task. While this metadata can be secured via encryption or steganography to prevent tampering, those techniques remain beyond the scope of this study.

\noindent \textbf{Authenticity of camera-captured images.} To our knowledge, \cite{access} is the first paper to highlight that even camera-captured images can contain non-authentic content. They address two scenarios: (1) when the neural ISP module operates as a black box, such as with third-party pretrained solutions, and (2) when camera engineers control the AI module’s architecture and training. The first scenario is more common and challenging, making it our focus.  While \cite{access} specifically tackles the detection of hallucinated pixels in both cases, our objective is to recover the authentic image.

\section{Methodology}
\label{sec:methodology}

Given the final output of the camera ISP $\mathbf{y}$ containing hallucinations, our objective is to recover the authentic image $\mathbf{x}$ before hallucinated content was introduced as a result of GenAI ISP operations. Typically, the initial stages of the ISP focus on accurate signal recovery and are thus less prone to hallucinations. Enhancement operations are usually applied at the end of the neural ISP processing chain~\cite{twostageisp,deepflexisp}, and these operations are more likely to introduce fake content. This is also typically true of super-resolution and low-light enhancement, the main tasks we focus on in this work, which reside at the end of the ISP pipeline. 
Our capture time processing is shown in Fig.~\ref{fig:method} (A). Following~\cite{access}, at capture time, we assume we have access to the intermediate image $\mathbf{x}$ on the ISP, which is the input to the hallucinatory AI model and the image we wish to recover post-capture. We first run the final image $\mathbf{y}\!\in\!\mathbb{R}^{H\!\times\!W\!\times\!3}$, the output of the AI model, through an encoder $\Phi$ that projects $\mathbf{y}$ to a $k$-dimensional embedding space $\Phi(\mathbf{y})\!=\!\mathbf{w}\!\in\!\mathbb{R}^{H\!\times\!W\!\times\!k}$. We then concatenate the $(x,y)$ coordinates to the encoder latents $\mathbf{w}$ and run it through a decoder MLP $\Theta$ to obtain the authentic image $\mathbf{\hat{x}}$. The decoder performs residual prediction, and we obtain $\mathbf{\hat{x}} = \mathbf{y} - \Theta([x,y,\mathbf{w}])$. The parameters of the encoder $\Phi$ and the MLP decoder $\Theta$ are stored as metadata accompanying the final output image $\mathbf{y}$. As shown in Fig.~\ref{fig:method}(B), the user can extract this metadata post-capture and recover the authentic image $\mathbf{\hat{x}} = \mathbf{y} -  \Theta([x,y,\Phi(\mathbf{y})])$.

\begin{figure}[t!]
\centering
\includegraphics[width=0.85\linewidth]{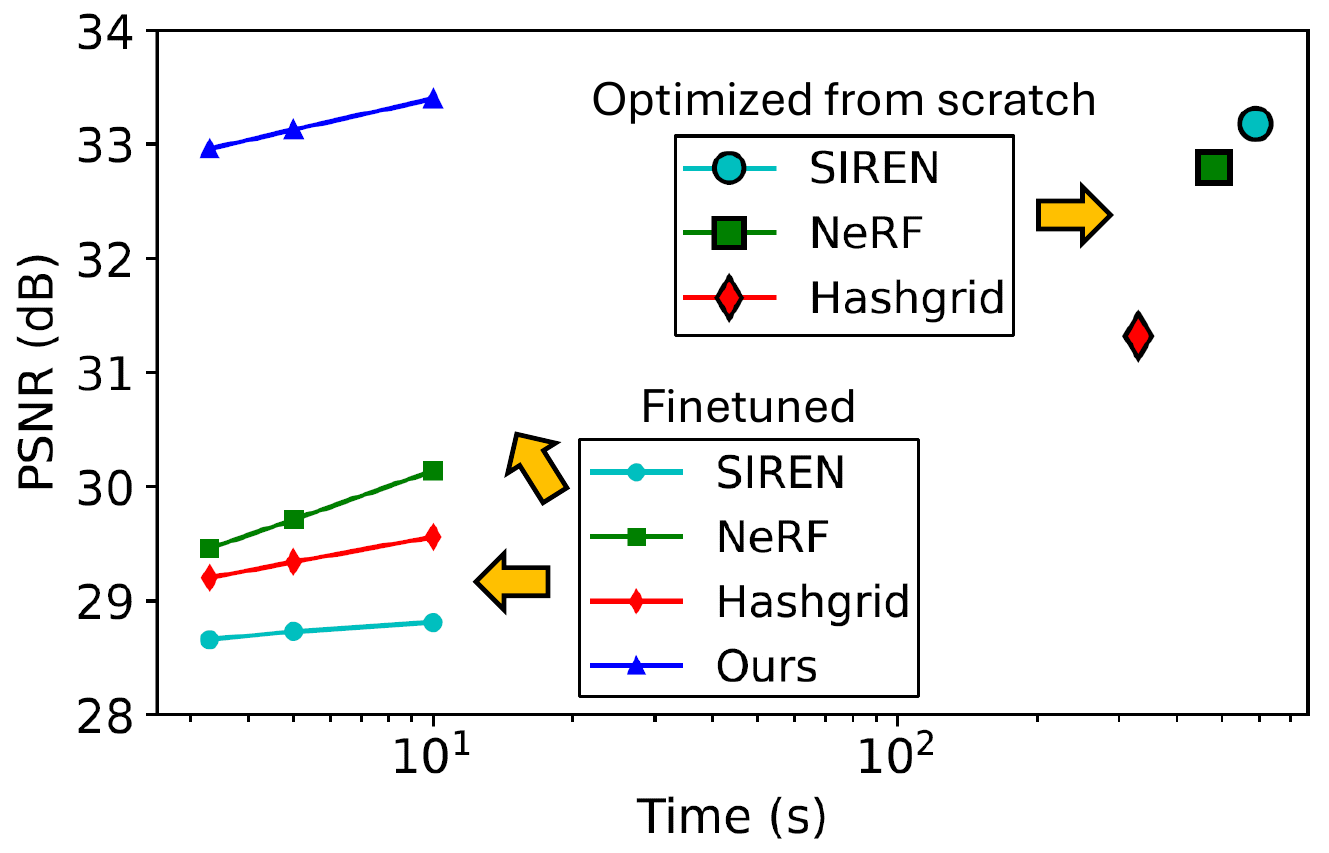}
\caption{Comparison of various methods against our proposed approach on training or finetuning time at image capture, versus PSNR (dB). Results are shown for the DIV2K~\cite{Agustsson_2017_CVPR_Workshops} dataset. While SIREN~\cite{siren}, NeRF~\cite{nerf} and hashgrid~\cite{hashgrid} perform well when optimized per image until convergence, training times are impractical. When finetuning under a constrained time budget, our method outperforms these competing approaches. 
\label{fig:psnr_time_plot}}
\end{figure}

\begin{figure*}[t!]
\centering
\includegraphics[width=0.975\linewidth]{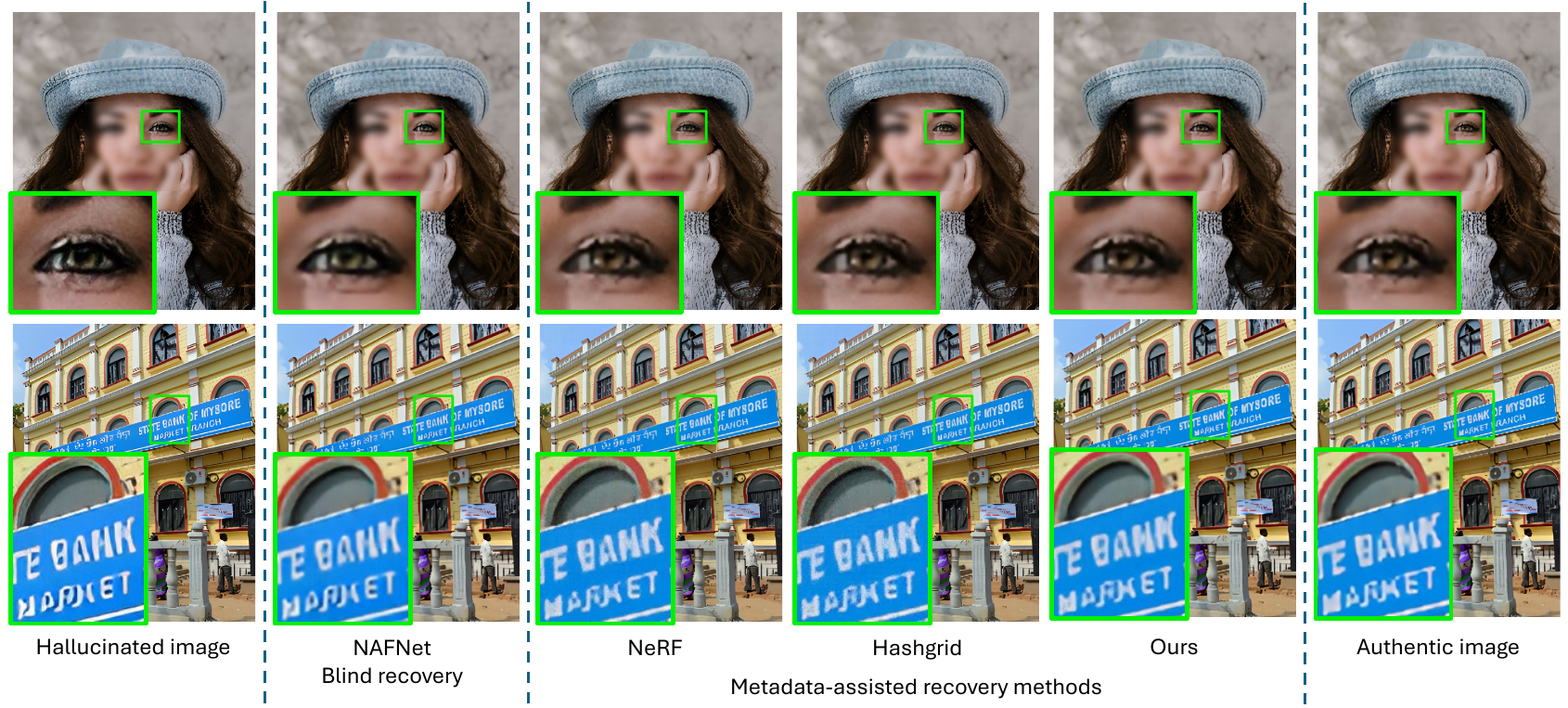}
\caption{Qualitative results for natural image super-resolution on the DIV2K~\cite{Agustsson_2017_CVPR_Workshops} dataset\protect\footnotemark. Inset shows zoomed-in region.}
\label{fig:esrgan_figure}
\vspace{-3mm}
\end{figure*}

The encoder and the MLP decoder are pretrained per modality on datasets of hallucinated and authentic image pairs. We use a reconstruction loss $||\mathbf{x}-\mathbf{\hat{x}}||_2^2$  at the output of the MLP $\Theta$. There is no intermediate supervisory loss for the encoder $\Phi$. Since we use only a reconstruction loss
and no perceptual or generative losses, our resultant image  $\mathbf{\hat{x}}$ does not risk introducing fake content. At capture time, the encoder weights are frozen, and the MLP decoder can be quickly finetuned for a small number of iterations for that image. We use the same loss function during finetuning. 

Our encoder architecture is inspired by NAFNet~\cite{nafnet}, a model that has demonstrated excellent performance on several image restoration tasks. We adopt a lightweight model for our encoder $\Phi$ with approximately 31.75 K parameters (127 KB) containing a single encoding block, middle block, and decoding block, each of the NAFNet architecture. We remove the long residual skip connection in the NAFNet since our encoder targets a latent representation, not image reconstruction. Our decoder MLP $\Theta$ has two hidden layers with 64 neurons in each layer and ReLU activation. The decoder has around 13 K parameters (53 KB). 

\begin{table}[t!]
\centering
\renewcommand{\arraystretch}{1.3}
\resizebox{\columnwidth}{!}{%
\begin{tabular}{ccccc}
\toprule
 & & \multicolumn{3}{c}{\textbf{PSNR (dB) on various datasets}} \\
 \cline{3-5}
\textbf{Methods}   & \begin{tabular}[c]{@{}c@{}} \textbf{Input} \\ \textbf{type} \end{tabular}  & \begin{tabular}[c]{@{}c@{}} \textbf{DIV2K} \\ \cite{Agustsson_2017_CVPR_Workshops} \end{tabular}    & \begin{tabular}[c]{@{}c@{}} \textbf{MARCO-} \\ \textbf{Net}~\cite{li2023marconet} \end{tabular} & \begin{tabular}[c]{@{}c@{}} \textbf{LOL} \\ \cite{Chen2018Retinex} \end{tabular} \\ 
\toprule
\multirow{2}{*}{SIREN~\cite{siren}}    & xyRGB & 28.75 & 27.56 & 34.87 \\
                          & xy & 19.57 & 20.60 & 25.43 \\ \midrule
\multirow{2}{*}{NeRF~\cite{nerf}}     & xyRGB & 29.46 & 26.63 & 36.24 \\
                          & xy    & 23.48 & 27.10   & 34.73  \\ \midrule
\multirow{2}{*}{Hashgrid~\cite{hashgrid}} & xyRGB & 29.20 & 30.32 & 35.65    \\
                          & xy    & 17.78 & 18.92  & 22.70   \\ \midrule
\begin{tabular}[c]{@{}c@{}} Blind \\ NAFNet~\cite{nafnet} \end{tabular}                     & -     & 32.25 & 27.22  & 23.04   \\ \midrule
\begin{tabular}[c]{@{}c@{}} Ours \end{tabular} & -                   & \textbf{32.96} & \textbf{31.26} & \textbf{36.34}    \\
\bottomrule    
\end{tabular}%
}
\caption{Results of our method against competing metadata-assisted MLP-based models, such as SIREN~\cite{siren}, NeRF~\cite{nerf} and hashgrid~\cite{hashgrid}, and blind recovery using a NAFNet~\cite{nafnet}. We evaluate the tasks of natural image SR, text SR, and low-light image enhancement on the DIV2K~\citep{Agustsson_2017_CVPR_Workshops}, MARCONet~\cite{li2023marconet}, and LOL~\cite{Chen2018Retinex} datasets, respectively. The best PSNR values are shown in bold.} 

\label{tab:pretraining_table}
\end{table}

\section{Experiments}
\label{sec:expts}

We evaluate our approach against competing methods on the two tasks of super-resolution and low-light image enhancement. For super-resolution (SR), we perform experiments on natural images as well as images of text. Hallucinations of text can easily alter image interpretation, and therefore, we study text SR separately as an important sub-class of general super-resolution. The encoder is pretrained per modality (e.g., one for natural image SR, one for text SR, one for low-light enhancement). The appropriate encoder is selected based on which AI ISP module was active during capture, information that is known to the camera.

\subsection{Datasets}
\label{subsec:datasets}
\noindent \textbf{Natural image SR.} We select the DIV2K~\cite{Agustsson_2017_CVPR_Workshops} dataset and a super-resolution factor of 4. We pick the RealESRGAN network~\cite{wang2021realesrgan} as a representative SR model for this experiment. In their paper, the authors first train the model using a pure reconstruction loss and then finetune it using an adversarial loss. The weights of both the reconstruction model and the final finetuned model are publicly available. Following~\cite{access}, we generate paired data where the output of the reconstruction model is the authentic image without hallucinations while the result of their final model finetuned with the GAN loss is the image with hallucinated content. The DIV2K dataset has 800 images for training and 100 images for validation. We split these 100 images further into 50 images for validation and 50 for testing.

\noindent \textbf{Text SR.} We use MARCONet~\cite{li2023marconet} as the text SR model for this experiment. As with natural image SR, we use a super-resolution factor of 4. We generate paired data where the images without hallucinations are obtained from the authors' synthetic text image generation pipeline, and the images with hallucinations are the output of the MARCONet model run on the former. Since MARCONet does not have a version trained with reconstruction loss only, we bicubically upsample the images without hallucinations by 4 to match the shape of the super-resolved images. We use the authors' official code to generate 2000 images, with 1600 images for training, 200 for validation, and 200 for testing.

\noindent \textbf{Low-light image enhancement.} We choose the LOL dataset~\cite{Chen2018Retinex} and the AutoDIR~\cite{jiang2023autodir} model. We use the low light images from the dataset as the authentic images with no hallucinations, and AutoDIR's output as the images with hallucinations. The LOL dataset has 500 images. We split these into 400 train, 50 validation, and 50 test images.

\subsection{Comparisons}
\label{subsec:comparisons}
\footnotetext{Note: Faces are blurred in the displayed figure for legal compliance.} 
We benchmark our approach against the following methods.

\noindent \textbf{(1) Implicit neural representations.} We compare against SIREN~\cite{siren} and NeRF~\cite{nerf} as the two most representative and accurate coordinate-based MLP frameworks. Both methods advocate a pre-defined and fixed encoding, as opposed to a learnable encoding proposed by our approach. SIREN applies a sine activation in all its layers, while NeRF opts for a frequency encoding in its first layer and ReLU activation in subsequent layers. For both methods, we test two variations -- the first where the $(x,y)$ coordinates are fed as input and a second where the $(x,y)$ coordinates and the RGB values of the hallucinated image are provided as input. For fair comparison, we use models with three hidden layers and 128 neurons per layer such that the models have a size of 204 KB that roughly matches the combined size of our encoder and MLP. As an upper-bound baseline comparison, we overfit the models per image from scratch till convergence for 100 K iterations. However, as noted earlier, this optimization takes order of minutes and is not a viable solution at capture time. We also compare against a more practical strategy similar to our proposed method. Here, we pretrain SIREN and NeRF on paired data under comparable settings as our method, and then finetune them at image capture for the same amount of time as our method. 

\begin{figure*}[t]
\centering
\includegraphics[width=0.975\linewidth]{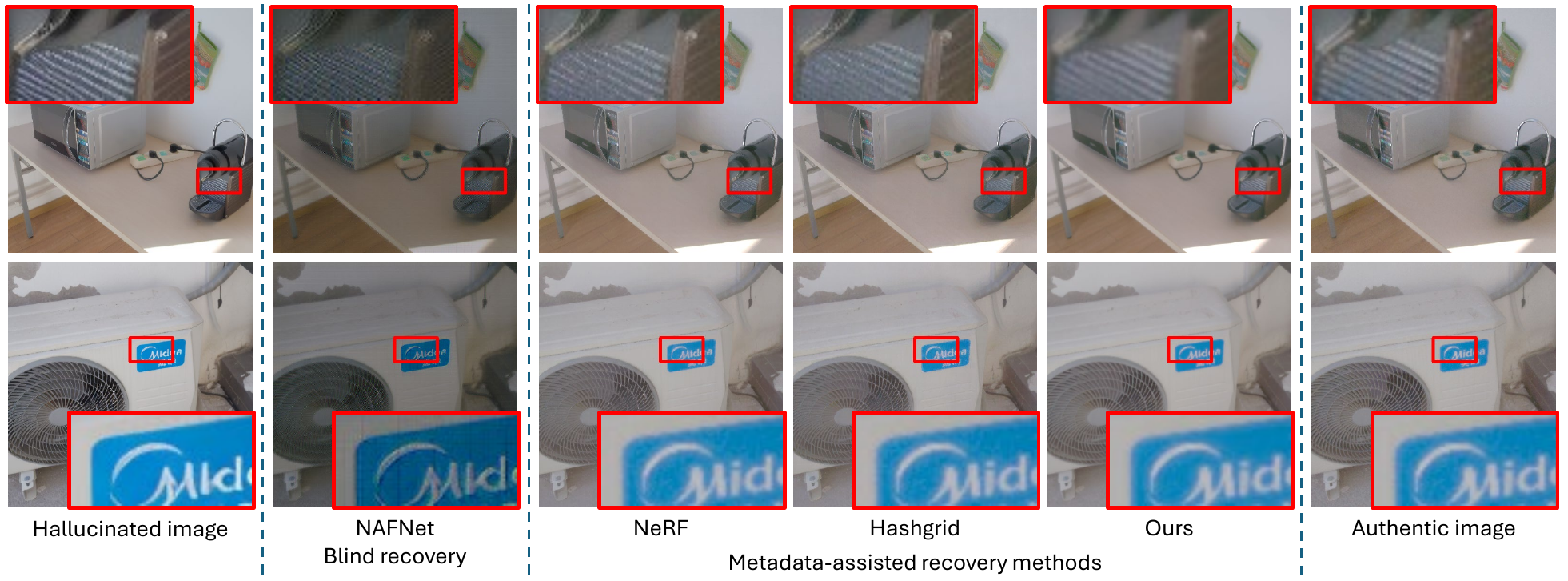}
\caption{Qualitative results for low-light image enhancement on the LOL~\cite{Chen2018Retinex} dataset. Inset shows zoomed-in region. For visualization, the authentic image and the results of all methods have been brightened to match the hallucinated input image.
\label{fig:lowlight_figure} 
}
\vspace{-3mm}
\end{figure*} 

\noindent \textbf{(2) MLP with a learned encoding.} We compare against the hashgrid approach in~\cite{hashgrid} as a representative method that, similar to our technique, combines a learnable embedding with an MLP. In particular, the entries of a hashtable are jointly optimized with the MLP, in~\cite{hashgrid}. For a fair comparison, we select the hyper-parameters of their method such that the embedding and the MLP are similar in size to our encoder and MLP, respectively. Please see the supplementary material for details. As before, we compare against two variants -- optimizing from scratch per image, and pretraining followed by finetuning. While training from scratch, both the hashtable and the MLP are optimized for 100 K iterations. When finetuning the pretrained model, we optimize only the MLP, similar to our method.

\noindent \textbf{(3) Blind image-to-image translation.} The comparisons in (1) and (2) assume the availability of image-specific metadata for recovery. We also compare against a no-metadata pixel-to-pixel regression approach. We use a large NAFNet model~\cite{nafnet} of size 64 MB following their original architecture. The training setup is described in the supplementary. However, as we shall show in our experiments, such a blind setup is sub-optimal, particularly for certain tasks, such as low-light enhancement, where there is an ambiguity when recovering the authentic image.

\subsection{Implementation details}
\label{subsec:implementation}
We implement our method using PyTorch. We use the Adam~\cite{Kingma:2014:Adam} optimizer for pretraining and finetuning. The parameters $\beta_1$ and $\beta_2$ are set to 0.9 and 0.999, respectively. We pretrain our encoder and MLP decoder for 50 K epochs with a batch size of 32. We pretrain a separate modality-specific encoder for each task, such as natural image SR, text SR, and low-light. The learning rate is initialized to $1e^{-4}$ and is reduced to $1e^{-5}$ after 40 K epochs. All MLP-based comparison methods are also pretrained for 50 K epochs, and other hyper-parameters are selected based on best performance or following the authors' recommendation in their papers. During finetuning, we use a learning rate of $1e^{-3}$ and a batch size of $2^{16}$ with random pixels drawn every iteration. For competing approaches, we use $1e^{-3}$ as the learning rate except for SIREN~\cite{siren} for which we found $1e^{-4}$ worked better. We finetune our method for 1000 iterations. This takes approximately 3 seconds on a V100 GPU with 32 GB RAM. While on-device optimization is required in practice, we limit the experiments to a desktop GPU due to the challenges of re-implementing all comparison methods on device, and for the sake of fair comparison. All competing methods were also evaluated on the same hardware. Other methods are also finetuned for the same amount of time as our method.

\subsection{Results}
\label{subsec:results}
Table~\ref{tab:pretraining_table} shows quantitative results of our method against competing approaches on the DIV2K~\cite{Agustsson_2017_CVPR_Workshops}, synthetic MARCONet~\cite{li2023marconet}, and LOL~\cite{Chen2018Retinex} datasets. For MLP-based methods, we try pretraining and then finetuning based on two types of input---$(x,y)$ and $(x,y,R,G,B)$. As expected, the models using just $(x,y)$ do poorly during pretraining, and thereby, find it difficult to achieve significant improvements within the limited finetuning time. Pretraining with $(x,y,R,G,B)$ leads to better performance.
Blind recovery works well on DIV2K~\cite{Agustsson_2017_CVPR_Workshops} for the natural image SR task but cannot achieve good results on MARCONet~\cite{li2023marconet} and LOL~\cite{Chen2018Retinex} because there is a one-to-many mapping in the solution space that the model cannot resolve. This is particularly evident in the low-light enhancement task where it is ambiguous for the model how dark the authentic low-light image was. Similarly, in the synthetic MARCONet data, images with different blur strengths map to the same hallucinated image. It can be seen from the results that our proposed approach outperforms competitors across all datasets.

The plot of Fig.~\ref{fig:psnr_time_plot} compares our method against competing approaches on time taken at image capture to finetune a pretrained model or to train a model from scratch, versus PSNR (dB). Results are shown for the DIV2K dataset~\citep{Agustsson_2017_CVPR_Workshops}. It can be observed that the MLP-based competitors SIREN~\cite{siren}, NeRF~\cite{nerf}, and hashgrid~\cite{hashgrid} perform well when trained from scratch per image (the markers on the top-right of the plot) but the training time is impractical since this optimization has to be done at capture time. Our method outperforms these competing strategies when adopting a more viable strategy of offline pretraining and quick capture time finetuning. The line plots show the results for various values of finetuning time where the three markers correspond to 3, 5 and 10 seconds, respectively. The $(x,y,R,G,B)$ input variant, which performs better than $(x,y)$ input, is shown for~\cite{siren,nerf,hashgrid}.

Fig.~\ref{fig:esrgan_figure} shows qualitative results from the DIV2K dataset~\cite{Agustsson_2017_CVPR_Workshops}. The first example highlights the type of hallucinations a GAN-based SR method~\cite{wang2021realesrgan} introduces to images containing faces---in this example, the color of the eye changes from brown to green. Blind approaches may struggle to recover the unhallucinated content in such situations. In the second image, it can be observed that the metadata-assisted methods, NeRF~\cite{nerf} and hashgrid~\cite{hashgrid}, exhibit noise-like artifacts around the window arch and the text that are not present in the hallucinated input image. Our recovery more closely matches the authentic image.

Fig.~\ref{fig:lowlight_figure} shows examples from the LOL dataset~\cite{Chen2018Retinex}. For visualization, we have scaled up the brightness of the low-light authentic image such that it matches the low-light-enhanced hallucinated image. The results of all methods have also been scaled up by the same factor for the purpose of visualization. Note that blind recovery cannot resolve this unknown scale factor. In the first result, our method reconstructs the striped pattern in the authentic image more accurately than competing approaches whose recovered texture more resembles the hallucinated input image. In the second example, it can be seen that the letter `i' has been hallucinated as an `l' as a result of low-light enhancement. Our method recovers the flipped character `i' while not introducing any artifacts in the homogeneous regions.

\begin{figure}[t]
\centering
\includegraphics[width=\linewidth]{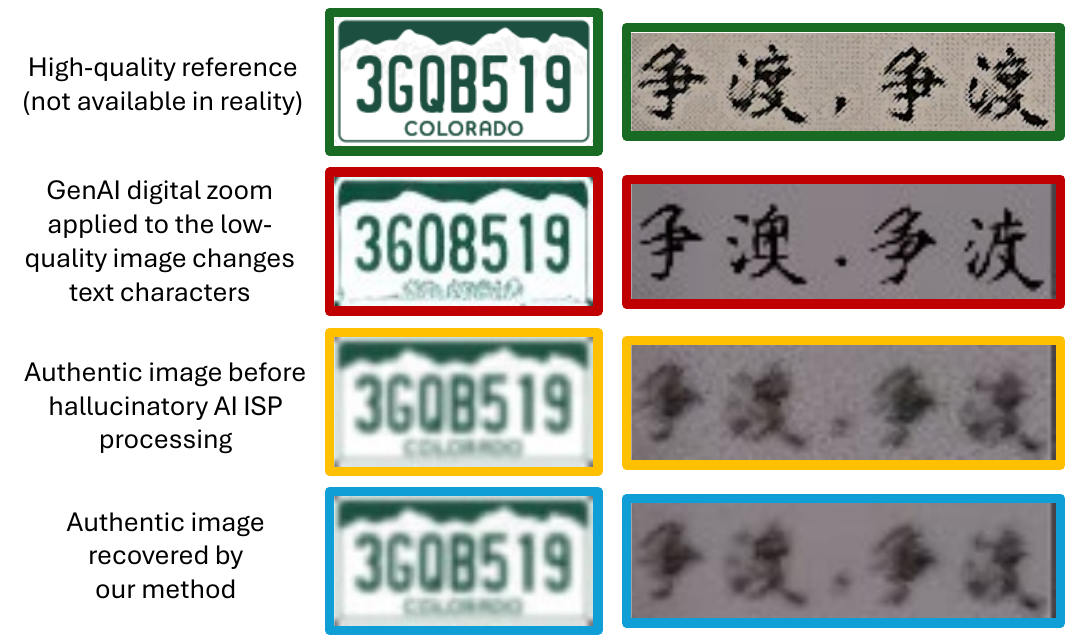}
\caption{Examples of hallucinations from GenAI-based text super-resolution. In the license plate image, `GQB' changed to `608', while the Chinese characters \begin{CJK*}{UTF8}{gbsn} 争渡，争渡 \end{CJK*} became \begin{CJK*}{UTF8}{gbsn} 争澳，争渡 \end{CJK*}. Our recovery closely matches the authentic image before super-resolution was applied.
\label{fig:last_figure} 
}
\vspace{-3mm}
\end{figure}

Intermediate images on the ISP, while accessible to camera manufacturers, are difficult to obtain publicly. Therefore, we use public datasets such as DIV2K~\cite{Agustsson_2017_CVPR_Workshops} and LOL~\cite{Chen2018Retinex} to quantitatively evaluate our recovery accuracy against other approaches. As previously mentioned, most hallucinations do not affect the interpretation of the scene (e.g., the first image in Fig.~\ref{fig:lowlight_figure}). However, when hallucinations alter the semantics, the consequences can be serious. For example, a GenAI super-resolved image directly outputted by a camera could potentially be used as evidence against the car's owner of the hallucinated license plate. Fig.~\ref{fig:last_figure} showcases two hand-picked examples for text SR that cause characters to change.

\subsection{Ablations}
\label{subsec:ablations}
We show ablations of our encoder and MLP architectures in Table~\ref{tab:ablations_table}. All results are reported on the DIV2K~\citep{Agustsson_2017_CVPR_Workshops} dataset. We first compare against a simple `base' encoder with approximately the same number of parameters as our modified NAFNet encoder. This base encoder contains cascaded blocks of convolution, batch normalization, and ReLU layers. 
Next, we vary the dimensionality of our encoder's embedding space for $k=0, 32, 128$ instead of our choice of 64, where $k=0$ refers to not using an encoder. We also ablate our MLP decoder by varying the number of neurons per hidden layer. 
Finally, we modified our encoder to have two blocks each of the NAFNet encoding block, middle block, and decoder block (denoted 2x NAFNet). We chose our modified NAFNet encoder with one block and an MLP with two hidden layers and 64 neurons as a good compromise between metadata size, accuracy, and finetuning time at image capture. Note that all ablations reported in Table~\ref{tab:ablations_table} are after finetuning for the same time budget as our method.

\begin{table}[t!]
\centering
\renewcommand{\arraystretch}{1.2}
\resizebox{\columnwidth}{!}{%
\begin{tabular}{cccc}
\toprule
\begin{tabular}[c]{@{}c@{}}\textbf{Encoder}\\ \textbf{architecture}\end{tabular} & \begin{tabular}[c]{@{}c@{}}\textbf{Decoder}\\ \textbf{architecture}\end{tabular} & \begin{tabular}[c]{@{}c@{}}\textbf{PSNR}\\ \textbf{(dB)}\end{tabular} & \begin{tabular}[c]{@{}c@{}}\textbf{Metadata}\\ \textbf{size (KB)}\end{tabular} \\ 
\toprule
\begin{tabular}[c]{@{}c@{}}Base encoder\end{tabular} & $k$=64,    MLP=64$\times$2 & 29.52 & 194 \\
\midrule
\multirow{6}{*}{\shortstack[c]{Modified\\ NAFNet}}    & 
$k$=0,    MLP=64$\times$2 & 28.61 & 53 \\
& $k$=32,    MLP=64$\times$2 & 32.88 & 153 \\
 & $k$=128,  MLP=64$\times$2           & 33.03 & 231\\
 & $k$=64,    MLP=32$\times$2          & 32.43 & 146 \\
 & $k$=64,    MLP=64$\times$2          & 32.96 &  180\\
 & $k$=64,    MLP=128$\times$2         & 33.45 & 293\\
 \midrule
 \begin{tabular}[c]{@{}c@{}}2x NAFNet \end{tabular} & $k$=64,    MLP=64$\times$2 & 33.50 & 248 \\
\bottomrule
\end{tabular}%
}
\caption{Ablations of different parameters of our proposed method. We vary the encoder architecture, the encoding dimension $k$, and the MLP size. 
Results are reported on the DIV2K dataset~\cite{Agustsson_2017_CVPR_Workshops}.}
\label{tab:ablations_table}
\vspace{-3mm}
\end{table}

\section{Conclusion}
\label{sec:conclusion}

In this work, we highlight the issue of authenticity in camera-captured images. Hallucinated content can appear even in images directly output by cameras due to the increasing integration of GenAI modules into modern camera hardware ISPs, particularly in smartphones. We advocate for a framework that allows users to access the unhallucinated image after capture.  

To this end, we propose storing the parameters of a lightweight MLP-based neural network as metadata alongside each camera-captured image. While our primary goal is to draw attention to the widely overlooked problem of hallucination in camera images, we also demonstrate through extensive experiments---on tasks such as super-resolution and low-light image enhancement---that our metadata-assisted approach offers improvements over alternative methods in terms of storage footprint, execution time, and accuracy.  

As laws and standards evolve around post-capture AI editing, camera image authenticity will face new legal and industry requirements. We hope our work provides a foundation for future research in this critical area.

\newcommand{\hbAppendixPrefix}{S}
\renewcommand{\thefigure}{\hbAppendixPrefix\arabic{figure}}
\setcounter{figure}{0}
\renewcommand{\thetable}{\hbAppendixPrefix\arabic{table}}
\setcounter{table}{0}
\renewcommand{\theequation}{\hbAppendixPrefix\arabic{equation}}
\setcounter{equation}{0}
\renewcommand{\thesection}{\hbAppendixPrefix\arabic{section}}
\setcounter{section}{0}

\clearpage

\twocolumn[{%
\centering
\Large \textbf{Supplementary Material}\\[1.5em]
}]

This supplementary material contains additional results and experimental details that could not be included in the main paper due to space constraints.

\section{Implementation details of comparison methods}
\label{sec:others}
As mentioned in Section~\ref{subsec:comparisons} of our main paper, we selected the hyper-parameters of the hashgrid~\cite{hashgrid} method such that their embedding and MLP are similar in size to our encoder and MLP, respectively. In particular, we chose the number of levels $L=16$, the number of feature dimensions per entry $F=4$, and the maximum entries per level (hash table size) $T=2^9$. The remaining hyper-parameters are the same as recommended by the authors. Under these settings, the encoder of the hashgrid~\cite{hashgrid} method has roughly 32 K parameters (128 KB). The MLP model is chosen to be similar to our method with two hidden layers and 64 neurons per layer, making the overall hashgrid model size 184 KB.

For the blind image-to-image translation experiment, we use a large NAFNet~\cite{nafnet} model with 36 blocks, same as the default configuration proposed by the authors. We change the default width from 32 to 24 to reduce overfitting. We use the Adam optimizer with $\beta_1$ = 0.9, $\beta_2$ = 0.999, weight decay 0, and train for 200 epochs with a learning rate of $1e^{-4}$. We train on patches of size $64\!\times\!64$ pixels using an $l_1$ loss and a batch size of 128. Because this is a no-metadata approach, we directly apply the trained models on the test set during inference and do not perform further finetuning on the test pairs.

\section{Additional qualitative results}
\label{sec:more_qualitative}
\begin{figure*}[t]
\centering
\includegraphics[width=\linewidth]{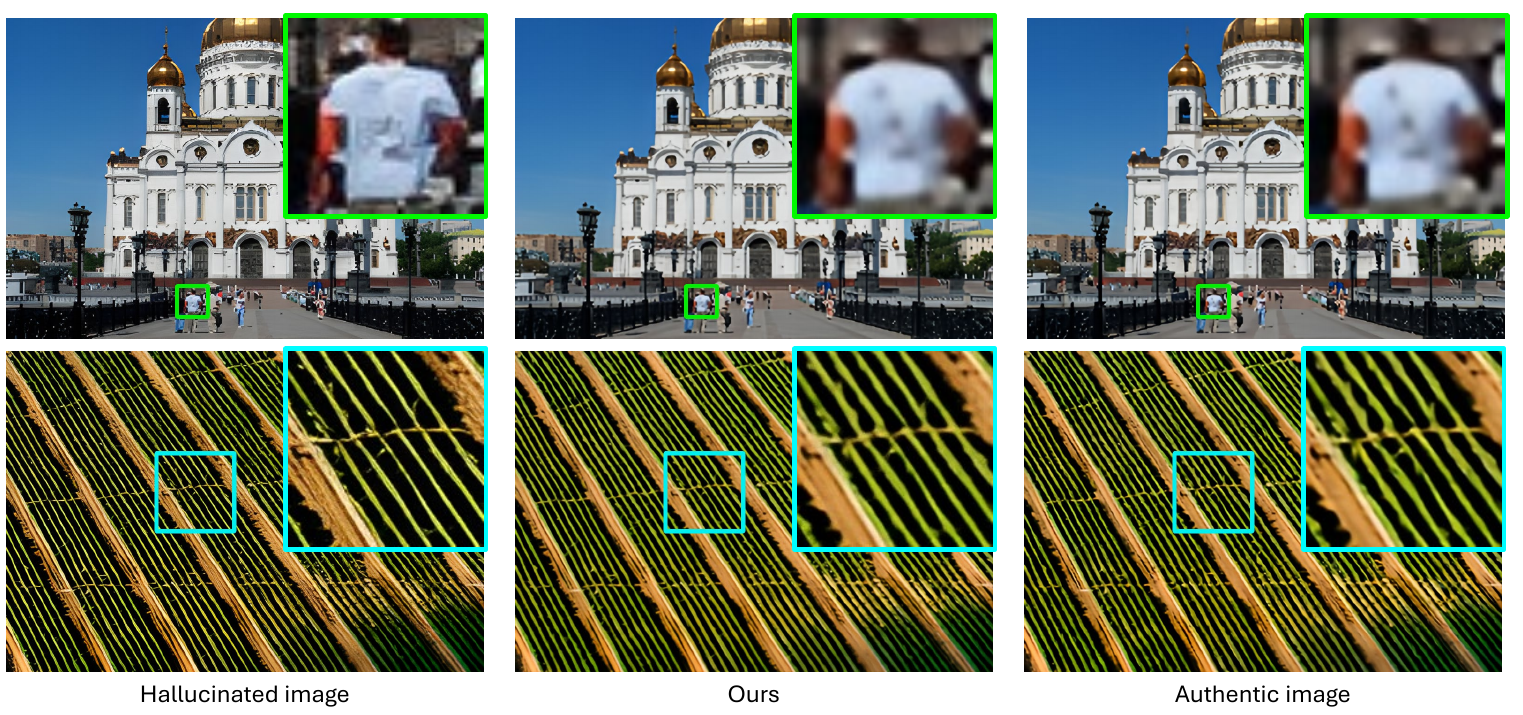}
\caption{Qualitative results for natural image super-resolution on the DIV2K~\cite{Agustsson_2017_CVPR_Workshops} dataset. Inset shows zoomed-in region.}
\label{fig:more_qualitative}
\end{figure*}

\begin{figure*}[t]
\centering
\includegraphics[width=\linewidth]
{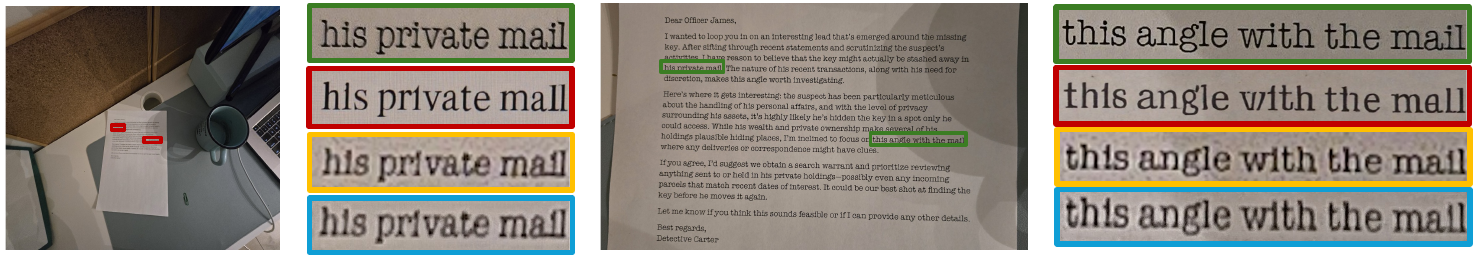}
\caption{Example of text SR on real images. In the first image, far-away text is enhanced by the AI ISP causing characters to flip (\textcolor{ForestGreen}{mail} to \textcolor{red}{mall}). The second image of the text region captured from a closer distance shows the actual text with no hallucinations (\textcolor{ForestGreen}{mail}). Such a reference image would not be available in practice. The authentic image (\textcolor{Dandelion}{yellow}) before hallucinatory text enhancement shows that the third letter of the last word is an ‘i’, which the AI model mistakenly super-resolved to an ‘l’. Our recovered authentic image (\textcolor{Cerulean}{blue}) can reverse this hallucination.}
\label{fig:marconet}
\end{figure*}

Fig.~\ref{fig:more_qualitative} shows additional qualitative results of our method on the DIV2K~\cite{Agustsson_2017_CVPR_Workshops} dataset described in Section~\ref{subsec:datasets} of our main paper. In the first example, the strap of the bag, that is missing in the hallucinated image, is correctly recovered by our method. In the second image of an aerial scene, the appearance of the crops is altered due to hallucinations. Our method is able to accurately recover the authentic image. Fig.~\ref{fig:marconet} shows a text SR example of our method applied to a real image.

\begin{table}[b]
\centering
\renewcommand{\arraystretch}{1.15}
\tiny
\resizebox{0.8\columnwidth}{!}{%
\begin{tabular}{ccc}
\toprule 
\textbf{Method} &
  \textbf{\begin{tabular}[c]{@{}c@{}}Metadata\\ size (KB)\end{tabular}} &
  \textbf{\begin{tabular}[c]{@{}c@{}}PSNR\\ (dB)\end{tabular}} \\ \toprule 
JPEG QF = 4                                                      & 181          & 27.44          \\
JPEG  QF = 15                                                      & 277          & 32.45          \\
JPEG QF = 30                                                     & 416          & 34.67          \\ \midrule
With binary mask  & 401          &    30.28            \\ \midrule
Error sampling  & 180          &    34.27            \\ \midrule
Ours & 180 & \textbf{35.12} \\ \bottomrule
\end{tabular}%
}
\caption{Comparison of our method against residual JPEG compression at different quality factors, saving a binary mask of hallucinated pixels as additional metadata, and finetuning based on weighted error sampling.}
\label{tab:metadata_schemes}
\end{table}

\section{Alternate metadata schemes}
\label{sec:metadata_schemes}
We propose to store the parameters of our encoder and our MLP decoder as metadata. Here, we compare this against directly saving the compressed {\it residual} between the hallucinated and authentic image pair as metadata. In particular, we applied lossy JPEG compression to the residual image with the quality factor (QF) chosen to yield a metadata overhead of around 180 KB, that is similar in size to our metadata, or higher. We perform this experiment on the DIV2K~\cite{Agustsson_2017_CVPR_Workshops} dataset. In the main paper, we had downsampled the images by a factor of four and then upsampled them back up by a factor of four, adhering to the protocol set by~\cite{Agustsson_2017_CVPR_Workshops}. For this comparison, we first downsample the images by a factor of two before super-resolving them by a factor of four such that the output images are approximately 12 MP resolution, which matches the resolution of most current smartphone cameras. We generate a paired dataset of hallucinated and authentic images using the RealESRGAN network~\cite{wang2021realesrgan}, following the same procedure described in Section~\ref{subsec:datasets} of our main paper. 

Table~\ref{tab:metadata_schemes} shows the result of residual JPEG compression at different quality factors against our approach. Our results are obtained using the same pretrained DIV2K encoder and MLP decoder from our experiments in the main paper. It can be seen that our approach performs significantly better at comparable metadata size (QF = 4). Even at more than double the metadata size (QF = 30), our result is more accurate. Also note that our metadata overhead is independent of image resolution, while the memory footprint of the residual image increases with image size.

We also compare against a mask-based metadata approach suggested by~\cite{access}. Following~\cite{access}, we threshold the residual between the hallucinated and authentic image pair to obtain a binary mask. Pixels that are flagged by the mask are considered hallucinated while the remaining pixels are authentic. This binary mask is saved as metadata, along with the encoder and MLP model weights. At inference, we finetune the MLP only on those pixels that are marked as hallucinated in the mask. 
The results are shown in Table~\ref{tab:metadata_schemes}. We see that this approach is not as accurate as our method. One drawback of this technique is that the mask adds considerable metadata overhead. As a proxy for the binary mask to flag fake pixels, we compare against a variant of our proposed method where we adapt the finetuning to focus on and correct those pixels that are more likely to be hallucinated. In particular, instead of randomly sampling pixels, we use a weighted sampling based on the per-pixel error map between the hallucinated and authentic image i.e., pixels with a higher error are sampled more frequently. However, we did not notice a significant improvement in the aggregate PSNR, as seen from the results in Table~\ref{tab:metadata_schemes}. Moreover, compared to random sampling, weighted error sampling is computationally more expensive and increases the time needed for finetuning.

\section{Modality-specific encoder}
\label{sec:encoder}

\begin{table}[b]
\centering
\renewcommand{\arraystretch}{1.25}
\resizebox{\columnwidth}{!}{%
\begin{tabular}{cccc}
\toprule
\textbf{}           & \multicolumn{3}{c}{\textbf{PSNR (dB) on various datasets}} \\ \cline{2-4} 
\textbf{Encoder}                                       & \textbf{DIV2K} & \textbf{MARCONet} & \textbf{LOL} \\ \midrule
Modality agnostic & 31.54              & 29.91             & 36.32             \\
Ours -- Modality specific & \textbf{32.96}          & \textbf{31.26}              & \textbf{36.34}   \\    \bottomrule
\end{tabular}%
}
\caption{Comparison between a modality-agnostic encoder and our modality-specific encoder.}
\label{tab:modality}
\end{table}

As mentioned in Section~\ref{subsec:implementation} of our main paper, we pretrain a separate modality-specific encoder jointly with the MLP for each task, such as natural image SR, text SR, and low-light image enhancement. In Table~\ref{tab:modality}, we compare against a generic encoder that is pretrained with the MLP on a mix of the training splits from all three datasets described in Section~\ref{subsec:datasets}. This modality-agnostic encoder is then evaluated on the test images from each dataset with the MLP decoder finetuning performed in an identical manner to our approach. Our results in the last row are reproduced from Table~\ref{tab:pretraining_table} of our main paper for ease of comparison. It can be observed that the modality-agnostic encoder performs worse than our modality-specific encoder with up to 1.5 dB drop in performance. This suggests that we need modality-specific pretraining to get the best performance.

\begin{table}[b]
\centering
\renewcommand{\arraystretch}{1.25}
\resizebox{\columnwidth}{!}{%
\begin{tabular}{ccc}
\toprule
\textbf{Description} & \textbf{Configuration} & \textbf{PSNR (dB)} \\
\toprule
\multirow{2}{*}{\begin{tabular}[c]{@{}c@{}}Different hidden\\ layers of MLP\end{tabular}} &
  MLP=64x1, 162 KB &
  32.73 \\
            & MLP=64x3, 196 KB & 32.95 \\ \midrule
Only latent & K=64, no $(x,y)$        & 32.75 \\ \midrule
\multirow{4}{*}{\begin{tabular}[c]{@{}c@{}}Random batch sampled\\ every @iteration\end{tabular}} &
  Sampling @100 &
  32.78 \\
            & Sampling @50          & 32.85 \\
            & Sampling @25          & 32.91 \\
            & Ours -- Sampling @1         & \textbf{32.96} \\ \midrule
\end{tabular}%
}
\caption{Additional ablations of our model architecture and training settings.}
\label{tab:ablation_supp}
\end{table}

We propose to save the encoder parameters as well as the MLP decoder parameters, with a combined size of 180 KB, as metadata along with each captured image. This makes the metadata self-contained and has the advantage that post-capture recovery requires just the image and its metadata. Alternately, since the encoder is frozen for a particular modality, only the MLP decoder weights, with a size of 53 KB, can be saved as metadata along with each image. However, this necessitates that the encoder's modality is known and its specific weights are available at recovery time through some other mechanism, such as retrieving it from a secure online repository. This may be less preferred in practice.

\section{Additional ablations}
\label{sec:more_ablations}

Table~\ref{tab:ablation_supp} shows additional ablations of our model architecture and training settings. All experiments are performed on the DIV2K~\cite{Agustsson_2017_CVPR_Workshops} dataset described in Section~\ref{subsec:datasets} of our main paper. Results reported are after finetuning. First, we vary the number of hidden layers of our MLP architecture. Increasing the number of layers to three from our choice of two increases the metadata size, but did not yield any notable difference in performance. We also tried not concatenating the $(x,y)$ coordinates as input to the MLP and only decoding from the $K$-dimensional latent feature from the encoder. Further, we show some ablations on the frequency of drawing random samples during finetuning, where @$p$ denotes that a random batch is drawn only every $p$ iterations. Sampling less frequently could speed up the finetuning time, however, as shown, it comes with decreased performance. 
Our proposed approach of sampling at every iteration, shown in the last row, gives the best results.

{
    \small
    \bibliographystyle{ieeenat_fullname}
    \bibliography{main}
}


\end{document}